\documentclass[conference]{IEEEtran}
\IEEEoverridecommandlockouts
\usepackage{cite}
\usepackage{amsmath,amssymb,amsfonts}
\usepackage{algorithmic}
\usepackage{graphicx}
\usepackage{textcomp}
\usepackage{xcolor}
\usepackage{booktabs} 
\usepackage[linesnumbered,ruled,vlined]{algorithm2e}
\usepackage{svg}
\usepackage{amsmath}
\usepackage{float}
\usepackage[font=small,labelfont=bf]{caption}
\usepackage[utf8]{inputenc}

\usepackage{multirow}
\usepackage{multicol}
\usepackage{makecell}
\usepackage[font=footnotesize]{subfig}
\usepackage{rotating}
\usepackage{enumitem}

\SetAlFnt{\small}

\def\BibTeX{{\rm B\kern-.05em{\sc i\kern-.025em b}\kern-.08em
    T\kern-.1667em\lower.7ex\hbox{E}\kern-.125emX}}
\begin{document}

\title{\huge SMATE: Semi-Supervised Spatio-Temporal Representation Learning on Multivariate Time Series}

\author{\IEEEauthorblockN{Jingwei Zuo, Karine Zeitouni and Yehia Taher}
\IEEEauthorblockA{DAVID Lab, University of Versailles, Université Paris-Saclay, Versailles, France \\
Email: $\{$jingwei.zuo, karine.zeitouni, yehia.taher$\}$@uvsq.fr}
}

\maketitle

\begin{abstract}
Learning from Multivariate Time Series (MTS) has attracted widespread attention in recent years. In particular, label shortage is a real challenge for the classification task on MTS, considering its complex dimensional and sequential data structure.
Unlike self-training and positive unlabeled learning that rely on distance-based classifiers,
in this paper, we propose SMATE, a novel semi-supervised model for learning the interpretable Spatio-Temporal representation from weakly labeled MTS.
We validate empirically the learned representation on 30 public datasets from the UEA MTS archive. We compare it with 13 state-of-the-art baseline methods for fully supervised tasks and four baselines for semi-supervised tasks. The results show the reliability and efficiency of our proposed method.
\end{abstract}


\section{Introduction}
Most Multivariate Time Series (MTS) data, such as sensor readings, are labeled during the data collection process. The post-labeling of MTS is much more costly than classic data (e.g., images, text, etc.) due to the low interpretability over the real-valued sequence, leading to a considerable constraint for MTS classification in real-life scenarios.

Weakly supervised learning becomes an alternative option to fully supervised algorithms. The previous studies on weak-label Time Series (TS) learning are usually based on self-training \cite{Wei2006Semi-supervisedClassification} or Positive Unlabeled Learning \cite{Nguyen2011PositiveClassification} with a carefully designed distance measure \cite{Chen2013DTW-D:Example} or stopping criterion \cite{RatanamahatanaStoppingClassification} to learn the pseudo-labels. Besides, they mostly focus on the Univariate Time Series (UTS) with the One-Nearest-Neighbor (1NN) classifier on raw data space, which is widely outpaced by today's advanced approaches \cite{Bagnall2020TheAdvances}, such as Deep Neural Networks (DNNs) \cite{Hao2020ASeries} or ensemble methods \cite{Bagnall2020TheAdvances}.  

From Univariate Time Series (UTS) to Multivariate Time Series (MTS), traditional methods usually combine the compact and effective features from different variables, such as Shapelet Ensemble \cite{Cetin2015ShapeletSeries,Mousheimish2017AutomaticProcessing}, global discriminative patterns \cite{Dorle2020LearningClassification}, or bag-of-patterns \cite{GokceBaydogan2015LearningClassification,Schafer2017MultivariateWEASEL+MUSE}. However, the predefined features usually fail to capture MTS essentials: the temporal dependency and the interactions of multiple variables (i.e., \textit{spatial interactions}\footnote{We use the term \textit{spatial} in this paper to represent the variable axis.}).
Recently, some deep learning-based methods were proposed to capture the MTS features with various network structures \cite{KarimMultivariateClassification,ZhengLNCSNetworks}, showing promising performance on MTS classification tasks. However, the above-mentioned methods are mostly fully supervised, and rarely consider the label shortage issue when building the MTS classifier.

The recent research turns to Representation Learning \cite{6472238} when handling weakly labeled MTS, which allows learning low dimensional embeddings in an unsupervised manner, such as using triplet loss \cite{Franceschi2019UnsupervisedSeries} to form the embedding space, then even an SVM classifier is powerful enough on the learned representation \cite{Franceschi2019UnsupervisedSeries}. However, existing techniques suffer from three major issues. First, the interactions between the MTS variables are generally computed on the entire 1-D series, ignoring the fact that the local spatial interactions at the sub-sequence level may evolve in the dynamic sequence, that is $spatial\ dynamics$. Second, the representation learned in a pure unsupervised manner depends mostly on the loss function selection. As no label information is utilized to learn the representation \cite{Franceschi2019UnsupervisedSeries}, there is a risk that it deviates from the true features, thus affecting the classifier performance. Third, they rather employ deep learning as a blind box and do not focus on the interpretability of the learned representation.  

Therefore, to handle both the MTS complex structure and the label shortage problem, we propose SMATE, \textbf{S}emi-supervised Spatio-temporal representation learning on \textbf{M}ultiv\textbf{A}riate \textbf{T}ime S\textbf{E}ries. An auto-encoder based structure allows mapping the MTS samples from raw features space $\mathcal{X}$ to low dimensional embedding space $\mathcal{H}$. A Spatial Modeling Block combined with a multi-layer convolutional network captures the spatial dynamics, whereas a GRU-based structure extracts the temporal dynamic features. Thereby, SMATE is capable of compressing the essential Spatio-temporal characteristics of MTS samples into low-dimensional embeddings. 
On top of this embedding space $\mathcal{H}$, we propose a semi-supervised three-step regularization process to bring the model to learn class-separable representations, where both the labeled and unlabeled samples contribute to the model's optimization. This regularization process comes with the capability of visualization at each step, making SMATE interpretable.

We summarize the paper's main contributions as follows:
\begin{itemize}
    \item \textbf{Spatio-temporal dynamic features in MTS:} We claim and demonstrate that the temporal dependency and the evolution of the spatial interactions (\textit{spatial dynamics)} are important for building a reliable MTS embedding.
    \item \textbf{Weak supervision of representation learning:} With limited labeled data, SMATE can learn reliable class-separable MTS representations for downstream tasks, such as MTS classification (MTSC).
    \item \textbf{Interpretable MTS embedding learning:} 
    SMATE allows for visual interpretability, not only from the class-separable representations but also in each step of the semi-supervised regularization process.
    \item \textbf{Extensive experiments on the MTS datasets:} The experiments are carried out on 30 MTS datasets from different application domains. The detailed evaluation with 13 supervised baselines and four semi-supervised work is provided, which shows the effectiveness and the efficiency of SMATE over the state of the art.
    
\end{itemize}

The rest of this paper starts with a review of the most related work. Then, we formulate the problems of the paper. Later, we present in detail our proposal SMATE, which is followed by the experiments on real-life datasets and the conclusion.

\section{Related Work}

\subsection{Multivariate Time Series Representation Learning}

\textbf{Definition 1 (multivariate time series):} A multivariate time series (MTS) $\textbf{x} \in \mathcal{R}^{T \times M}$ is a sequence of real-valued vectors: $\mathbf{x}$=($x_1$, $x_2$, ..., $x_i$, ..., $x_T$), where $x_i \in \mathcal{R}^{M}$, $M$ is the variable number. When $M=1$, we call it \textit{univariate time series} (UTS).

A natural way to learn MTS representation is to extend the approaches developed earlier on UTS \cite{Zuo2019ExploringSE4TeC,Zuo2019IncrementalStream,Zuo2019ISETS}. For instance, \cite{Cetin2015ShapeletSeries} combines Shapelet representation from different variables to build an ensemble-like learner. Similarly, SMTS \cite{GokceBaydogan2015LearningClassification} and WEASEL+MUSE  \cite{Schafer2017MultivariateWEASEL+MUSE} adopt the symbolic features with the Bag-of-Patterns concept to model the variable's relationship.

Different from the interpretable feature-based representations, various deep learning models are proposed to capture the spatial interactions in MTS.
MC-DCNN \cite{ZhengLNCSNetworks} extracts firstly 1D-CNN features from each variable, then combines them with a Fully Connected (FC) Layer. Whereas the authors in \cite{IJCAI1510710} abandon the combination strategy but apply directly 1D-CNN to all variables. The 2D-CNN features with the cross-attention mechanism in CA-SFCN \cite{Hao2020ASeries} enhanced the dependencies captured by 1D-CNN on both temporal and spatial axes. 
Last but not least, the hybrid LSTM-CNN structure is capable of extracting both local and long-term features. Various work, such as the Squeeze-and-Excitation block in MLSTM-FCN \cite{KarimMultivariateClassification} or the multi-view learning-like module in TapNet \cite{ZhangTapNet:Network}, enhanced the hybrid structure via modeling the spatial interactions. 
However, the interactions are generally computed at the sequence level, ignoring the fact that the local spatial interactions at the sub-sequence level may evolve in the dynamic sequence, i.e., \textit{spatial dynamics}.
Moreover, they are all fully supervised, requiring huge labels during the training process. Also, the learned representations are result-oriented (e.g., pursuing higher accuracy), with less focus on the interpretability, considered by the data mining community.

\subsection{Semi-supervised Learning on Time Series}
The pioneering work \cite{Wei2006Semi-supervisedClassification,Chen2013DTW-D:Example} on Semi-supervised TS Learning are based on self-training or Positive Unlabeled Learning \cite{Nguyen2011PositiveClassification} with the Nearest-Neighbor (1NN) classifier and a carefully designed distance, such as DTW \cite{Wei2006Semi-supervisedClassification} or DTW-D \cite{Chen2013DTW-D:Example}, and optimized stopping criterion \cite{RatanamahatanaStoppingClassification} for importing the pseudo-labels. Those work are followed by \cite{JawedSelf-SupervisedClassification} for wider contexts. Though not mentioned in their papers, the self-training framework is extensible from the UTS to the MTS setting by using an adapted distance \cite{Shokoohi-Yekta2017GeneralizingAccess}, such as DTW$_{D}$ or DTW$_{A}$. However, under more complex scenarios nowadays, such as 30 UEA MTS datasets \cite{Hoang2018The2018} collected from different domains, the distance-based classifiers show limited performance \cite{Bagnall2020TheAdvances}.

Learning MTS representations \cite{6472238} in a weakly supervised setting draws much attention recently. Unsupervised Scalable Representation Learning (USRL) described in \cite{Franceschi2019UnsupervisedSeries} combines causal dilated convolutions with triplet loss for contrastive learning.
Similarly, authors in \cite{JawedSelf-SupervisedClassification} adopt Multi-Task Learning (MTL) to learn the self-supervised UTS features from an auxiliary forecasting task. The recent work Semi-TapNet \cite{ZhangTapNet:Network} proposes an Attentional Prototype Network to learn from the unlabeled samples.
However, the above-mentioned approaches do not explore thoroughly both the labeled samples and the \textit{Spatio-temporal dynamics} in MTS. We show in Table \ref{method_comparison} the comparison of the methods for learning MTS representation.
\begin{table}[htbp]
\centering
\caption{Existing methods for learning MTS Representation}
\label{method_comparison}
\scalebox{0.8}{
\begin{tabular}{|p{3.5cm} | c | c| c| c| c|c|c|c|c|}
\hline

  &\rotatebox{90}{SMTS} &\rotatebox{90}{MUSE} &\rotatebox{90}{Shapelet} & \rotatebox{90}{USRL} & \rotatebox{90}{TapNet} &\rotatebox{90}{\parbox{0.7cm}{MLSTM FCN}} & \rotatebox{90}{\parbox{0.7cm}{CA-SFCN}} & \rotatebox{90}{\parbox{0.7cm}{1NN-DTW}} & \rotatebox{90}{\textbf{SMATE}}\\
\hline
Temporal Dynamics & - & - & - &$\checkmark$ &$\checkmark$ &$\checkmark$ &$\checkmark$  &- &$\checkmark$\\

Spatial Dynamics & - & - & - &- &- &- &$\checkmark$ &- &$\checkmark$\\

Interpretable Representation & $\checkmark$ & $\checkmark$  & $\checkmark$ &- &$\checkmark$  &- &- &- &$\checkmark$\\

Label Shortage & - & - & - &$\checkmark$ &$\checkmark$ &- &- &$\checkmark$ &$\checkmark$\\
\hline
\end{tabular}}
\vspace{-1em}
\end{table}

\section{Problem Formulation}
In this section, we formulate the \textit{Spatio-temporal dynamics} learning and semi-supervised classification problems for MTS. 
Table \ref{Math_notation} summarizes the notations used in the paper.

\begin{table}[htbp]
\centering
\vspace{-1em}
\caption{\label{Math_notation}Notation}
\scalebox{0.8}{
\begin{tabular}{|p{3cm}| l |}
\hline
\small
Notation & Description\\
\hline
\hline
$\mathcal{D}, \mathcal{D}_{l}, \mathcal{D}_{u}$ & dataset, labeled portion, unlabeled portion\\
$T$, $M$, $N$ & MTS length, variable numbers, dataset size\\
$L, D$& embedding length, embedding dimension size\\
$m, \mathcal{P}$ & temporal window size, embedding pool size\\
$\mathbf{x}$, $\mathbf{h}$ &MTS instance, latent embedding\\
$\textbf{s}$       & variable/spatial interaction\\ 
$\theta$ & general parameters to be optimized\\

\hline
\end{tabular}}
\vspace{-2em}
\end{table}

\subsection{Spatio-temporal Representation for MTS}


The Spatio-temporal modeling of MTS requires considering both the temporal dependency $p(x_{{t}'} | x_{t})$ ($t'>t$) and the spatial interactions between the variables. Previous studies \cite{ZhangTapNet:Network,KarimMultivariateClassification} usually consider the spatial interactions at the sequence level: $\mathbf{s} = \{\textbf{x}^{i}\Join\textbf{x}^{j}\} \in \mathcal{R}^{M}$, where $\textbf{x}^{i}, \textbf{x}^{j} \in \mathcal{R}^{T \times 1}$, $\Join$ indicates the interactions between the variables. 
However, the local spatial interactions at the subsequence level $s_t = \{\textbf{x}_{{t-m/2}, {t+m/2}}^{i}\Join\textbf{x}_{{t-m/2}, {t+m/2}}^{j}\} \in \mathcal{R}^{M}$ may evolve in the dynamic sequence, where \textit{m} is the window size. To illustrate, given the system status at time $t$ in MTS, it is not only decided by the local value $x_{{t}} \in \mathcal{R}^M$ given a temporal status,  but also by its neighbor values $\left [ x_{{t-m/2}}:x_{{t+m/2}}\right ] \in \mathcal{R}^{M \times m}$, which brings a spatial correlation matrix on temporal neighbors and spatial variables given a spatial status $s_{t} \in \mathcal{R}^M$.

Therefore, given a sample $\mathbf{x} \in \mathcal{R}^{T \times M}$ in raw space $\mathcal{X}$, the Spatio-temporal representation learning for MTS is to learn a low-dimensional representation $\mathbf{h} \in \mathcal{R}^{L \times D}$ on embedding space $\mathcal{H}$, integrating both temporal dynamic $p(x_{{t}'} | x_{t})$ and spatial dynamic $p(s_{{t}'} | s_{t})$. The item \textit{dynamic} refers to the unstable system status within the evolving multivariate sequential data. 

\subsection{Semi-Supervised Learning on MTS}

\textbf{Definition 2 (weak-label MTS dataset):} A weakly labeled MTS dataset $\mathcal{D} = \{\mathcal{D}_{l}, \mathcal{D}_{u}\}$ contains both labeled and unlabeled MTS samples: 
\begin{equation}
\nonumber
\small
\begin{gathered}
    \mathcal{D}_{l}= \{\textbf{x}_{i},y_i\}_{i=1}^{N * r},\;
    \mathcal{D}_{u} =\{\hat{\textbf{x}}_{i}\}_{i=1}^{N * (1-r)}
\end{gathered}
\end{equation}
$r$ ($0 \leq r \leq 1$) indicates the \textit{ratio} of the labeled samples in $\mathcal{D}$ of size \textit{N}, $y_i$ is the annotation of the labeled instance $\mathbf{x}_i$.

The semi-supervised MTS learning aims at training a classifier to predict successfully the label of a testing MTS sample, adopting the supervised training from $\mathcal{D}_{l}$ and further unsupervised adjustment/optimization from $\mathcal{D}_{u}$.

\section{Proposal: SMATE}
In this section, we introduce SMATE, which captures the essential characteristics of MTS and allows learning an interpretable representation space in a semi-supervised manner.

\subsection{Global Structure of SMATE}

\begin{figure*}[htbp]
\centering
\hspace{-7em}
\begin{minipage}[t]{0.7\textwidth}
\centering
\includegraphics[width=11cm, height=5cm]{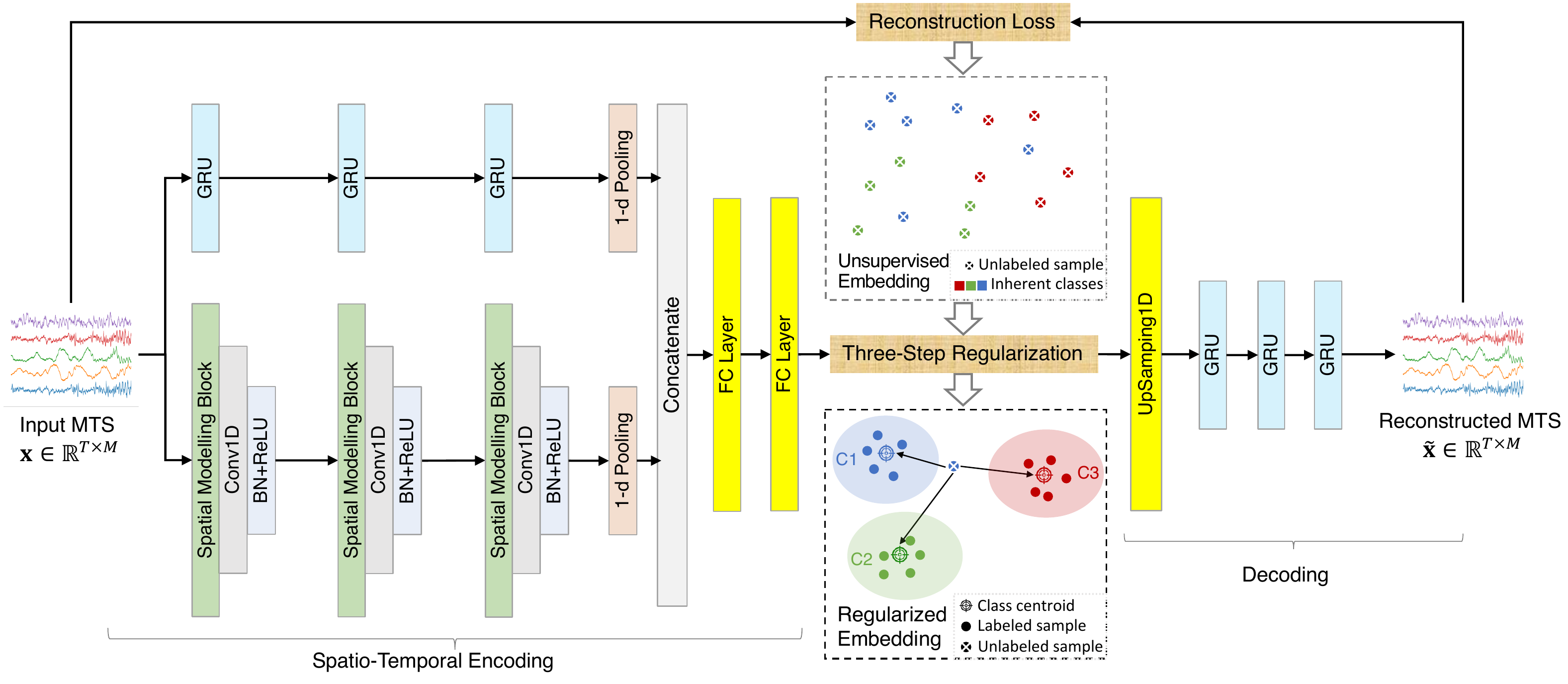}
\caption{Model Structure of SMATE}
\label{Model Structure of SMATE}
\end{minipage}
\hspace{-2em}
\begin{minipage}[t]{0.3\textwidth}
\centering
\includegraphics[width=7cm, height=4cm]{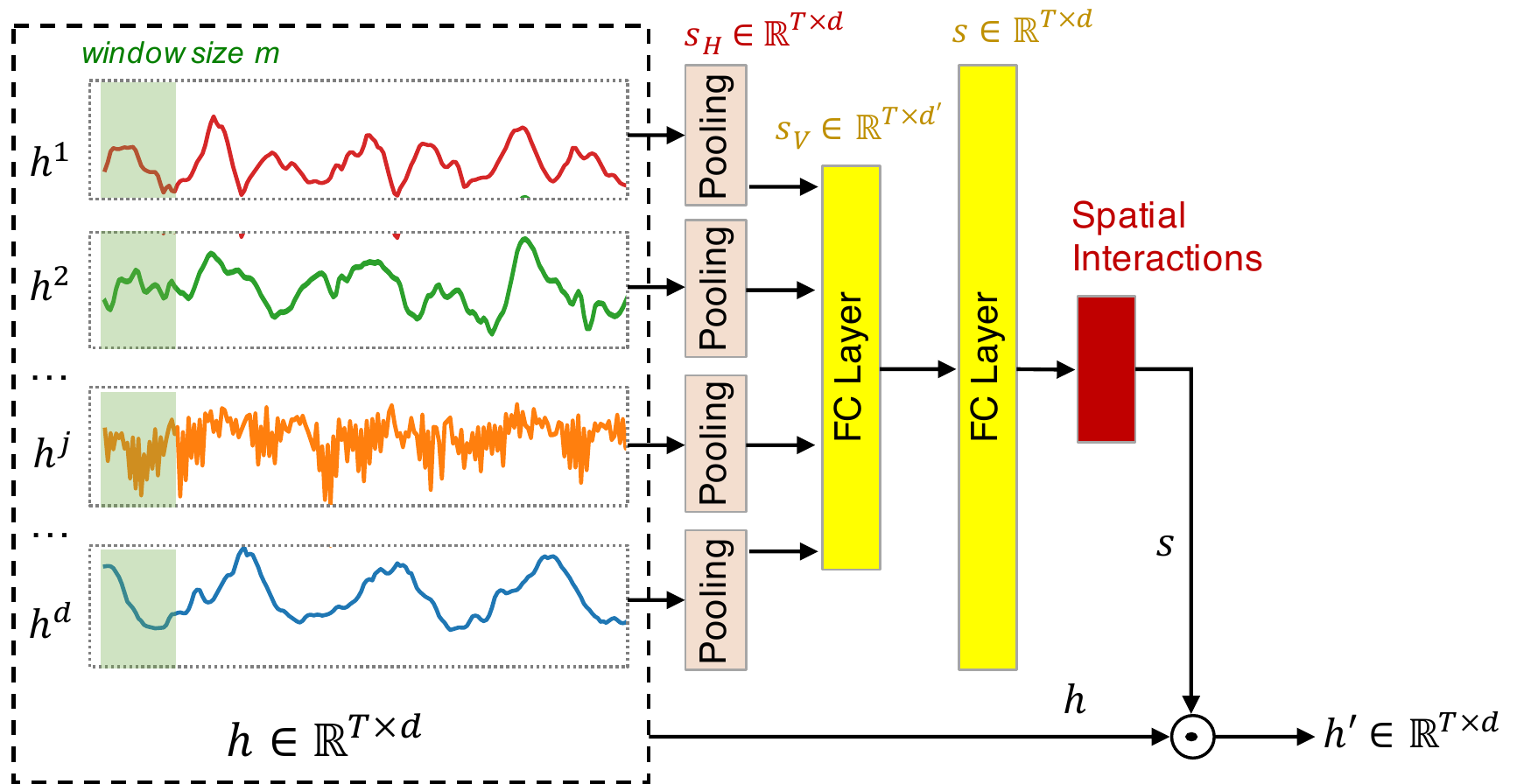}
\caption{The Spatial Modeling Block(SMB)}
\label{SpatialModelingBlock}
\end{minipage}
\vspace{-2em}
\end{figure*}

SMATE is based on an \textit{asymmetric auto-encoder} structure, integrating three key components: Spatio-temporal dynamic encoder, sequential decoder, and semi-supervised three-step regularization of the embedding space. 

As shown in Figure \ref{Model Structure of SMATE}, a two-channel encoder explores both spatial and temporal dynamics and embeds the input MTS into a low-dimensional representation space, where the embedded samples are sparsely distributed with the reconstruction-based optimization. On the unsupervised embedding space, a three-step regularization is applied to learn class-separable embeddings. The class centroids are regularized by labeled and unlabeled samples, showing interpretability over the representation space. Finally, the model is jointly optimized by the reconstruction objective and the regularization objective.  

\subsection{Spatial Modeling Block (SMB)}
Firstly, we introduce a novel module, Spatial Modeling Block (SMB), to capture the spatial interactions at subsequence levels. As shown in Figure \ref{SpatialModelingBlock}, SMB takes as input an MTS representation $h = \{h_i\}_{i=1}^{T} \in \mathcal{R}^{T\times d}$ ($d=M$ for the first block in spatial encoding channel), followed by a one-dimensional average pooling layer on each variable $h^j \in \mathcal{R}^{T\times1}$, encoding the temporal neighbors into the horizontal system status $s_{H}(i)$=$avg(\left [ h_{{i-m/2}}:h_{{i+m/2}}\right ])$, where $i$ is the time tick, $m$ is the window size.
Then the Fully Connected (FC) layers allow firstly the interaction of the horizontal system status $s_H$ in the vertical direction via a low-dimensional compression $s_V \in \mathcal{R}^{T\times d'}$, then remapping it to the initial data space to decide the spatial interaction weights at each one-dimensional segment.
We define the spatial interactions $\mathbf{s} = \{s_i\}_{i=1}^{T}$, where $s_i \in \mathcal{R}^d$, representing the interaction weights for the vector $h_i \in \mathcal{R}^d$. The output of SMB is described by $h'$=$h \odot  \mathbf{s}$, with the calibrated weights for each 1-D TS segment, where $\odot$ is the element-wise multiplication.

\subsection{Spatio-Temporal Encoding on MTS}
Given $\mathbf{x} \in \mathcal{R}^{T \times M} $, the low-dimensional representation $\mathbf{h} \in \mathcal{R}^{L \times D} $ embeds the Spatio-temporal features of $\mathbf{x}$ by a neural network-based function $f_{\theta}(\mathbf{x})$. The low-dimensional embedding brings dramatic improvement on both the efficiency and accuracy for classification tasks \cite{ZhangTapNet:Network}, as the classifier is not impaired by the redundant information in raw data.

As shown in Figure \ref{Model Structure of SMATE}, for the temporal channel, among different variants of the recurrent neural networks (RNN), we specifically consider Gated Recurrent Units (GRUs) that mitigate the vanishing gradient problem. An update gate $z_t$ and a reset gate $r_t$ control the hidden state $h_{t} \in \mathcal{R}^{d_{g}}$ with the observation $x_t \in \mathcal{R}^{M}$ and the previous hidden state $h_{t-1}\in \mathcal{R}^{d_{g}}$, where $d_{g}$ is the output dimension of GRUs. The update functions are as follows:
\vspace{-0.5em}
\begin{equation}
\small
r_t = \sigma(W_r x_t  + U_r h_{t-1} + b_r)
\vspace{-0.5em}
\end{equation}
\begin{equation}
\small
z_t = \sigma(W_z x_t  + U_z h_{t-1} + b_z) 
\vspace{-0.5em}
\end{equation}
\begin{equation}
\small
h_t = (1-z_t)\odot h_{t-1}  + z_t \odot \tanh (W_h x_t  + U_h(h_{t-1} \odot r_t) + b_h)
\vspace{-0.5em}
\end{equation}
where $W_x, U_x, b_x$ (\textit{$x\in[r,z,h]$}) are model parameters, $\sigma$ is the sigmoid function, $\odot$ is the element-wise multiplication. Three GRUs are cascaded with a 1-D pooling layer to output $\mathbf{h}(\mathbf{T}) \in \mathcal{R}^{L \times d_{g}}$, where $L$=$T/\mathcal{P}$, $\mathcal{P}$ is the pool/sampling size. 

For the spatial channel, we define the convolutional module:
\vspace{-0.5em}
\begin{equation}
\small
\mathbf{h}'(l) = SMB(\mathbf{h}(l))
\vspace{-0.5em}
\end{equation}
\begin{equation}
\small
\mathbf{h}(l+1) = ReLU(BN( W \otimes \mathbf{h}'(l)  + b)
\vspace{-0.5em}
\end{equation}
where $l$ ($0 \leq l < 3$) is the module number, $\mathbf{h}(0)$=$\mathbf{x}\in \mathcal{R}^{T \times M}$, $\mathbf{h}(l)\in \mathcal{R}^{T \times d_c}$, $d_c$ is the filter number, $W$ is the 1-D convolutional kernel of size $m$, $\otimes$ is the convolution operator.
Within each of the three modules, the SMB firstly calibrates the interaction weights for each 1-D segment and outputs $\mathbf{h}' \in \mathcal{R}^{T\times d}$. 
Then a 1-D convolutional layer concatenated with Batch Normalization and Rectified Linear Units (ReLU) is deployed.
Similar to the temporal channel, a 1-D pooling layer is applied after the last convolutional block to output $\mathbf{h}(\mathbf{S}) \in \mathcal{R}^{L \times d_{c}}$.
Finally, we output the combined spatial and temporal features $h_{concat}$ = $concat(\mathbf{h}(\mathbf{T})$, $\mathbf{h}(\mathbf{S})$) $\in \mathcal{R}^{L \times (d_g + d_c)}$ and apply two FC layers to get the Spatio-temporal embedding $\mathbf{h} \in \mathcal{R}^{L \times \mathbf{D}}$. The matrix representation captures the Spatio-temporal features and facilitates the MTS restoration.

\subsection{Joint Model Optimization}
As shown in Figure \ref{Model Structure of SMATE}, since the representation learned via an autoencoder-based structure generally has a sparse distribution of class-specific samples \cite{6472238}, the unsupervised training derived from the reconstruction objective does not consider thoroughly the inner relation between class-specific samples but focus on the restoration performance from the embeddings. To address the issue, we propose a joint model optimization that integrates the temporal reconstruction and the three-step regularization objectives. Specifically, the joint optimization combines both the labeled and unlabeled samples to learn the class-specific clusters on the embedding space. 

Firstly, we define the \textit{temporal reconstruction loss} as:
\begin{equation}
    L_R = \sum\nolimits_{t} \left \| x_{t} - \tilde{x}_t \right \|_2 
\end{equation}

where $x_{t},\tilde{x}_{t} \in \mathcal{R}^M$, corresponding to the observations in the raw and reconstructed MTS instances $\mathbf{x}$ and $\tilde{\mathbf{x}}$.

Then, the three-step regularization approaches the embeddings within the class-specific clusters to the virtual class centroids, which are trained progressively. 

\noindent\textbf{\textit{Step 1} Supervised Centroids Initialization}: 
The class centroids are initialized by the class-specific embeddings. Given the labeled training set $\mathcal{D}_{l} = \{X^{k} \}_{k=1}^{K}$ where $K$ is the class number, $X^{k}\in \mathcal{R}^{N_k \times T \times M}$ is a sample collection of class $k$, $N_k$ is the sample number of class $k$. Then the embedding set $H^{k}$=$f_{\theta}(X^k) \in \mathcal{R}^{N_k \times L \times D}$ initializes the class centroid $\mathbf{c}_k$ by:
\begin{equation}
    \mathbf{c}_k = mean(H^{k}), \quad \mathbf{c}_k \in \mathcal{R}^{L \times D}
\end{equation}

\noindent\textbf{\textit{Step 2} Supervised Centroids Adjustment}: Once the centroids are initialized, we can make the supervised adjustment since the distance-based class probability allows to assess the contribution of individual samples on the centroid's decision. In other words, the centroid $\mathbf{c}_k$ is affected by larger contribution weights brought by nearby samples of class $k$. Let $\mathbf{x}_i^k \in \mathcal{R}^{T \times M}$ be a time series of class $k$, we define the weight of $\mathbf{x}_i^k$ to $\mathbf{c}_k$ as the inverse Euclidean Distance (ED) between the embedding $\mathbf{h}_i^k=f_{\theta}(\mathbf{x}_{i}^k)\in \mathcal{R}^{L\times D}$ and the centroid $\mathbf{c}_k$:
\begin{equation}
    W_{k, i} = 1 - \frac{ED(\mathbf{h}_i^k, \mathbf{c}_k)}{\sum_{j=1}^{K} ED(\mathbf{h}_i^k, \mathbf{c}_j)}
\end{equation}
Then the class centroid $\mathbf{c}_k$ can be adjusted accordingly by the labeled samples within the class-specific cluster:
\begin{equation}
    \mathbf{c}_k = \sum\nolimits_{i=1}^{N_k} W_{k,i} \cdot \mathbf{h}_i^{k}, \quad \mathbf{h}_i^{k} \in H^k
    \label{Class centroids}
\end{equation}

\noindent\textbf{\textit{Step 3} Unsupervised Centroids Adjustment}: Apart from the reconstruction-based optimisation, the unlabeled sample $\hat{\mathbf{x}}_{i}$ in $\mathcal{D}_{u} = \{\hat{\mathbf{x}}_{i} \}_{i=1}^{N * (1-r)}$ is capable of adjusting the centroid $c_k$ via the propagated label from the distance-based class probability: 
\begin{equation}
    \hat{p}_{\theta}(y=k | \hat{\mathbf{x}}_i) = 1 - \frac{ED(f_{\theta}(\hat{\mathbf{x}}_i), \mathbf{c}_k)}{\sum_{j=1}^{K} ED(f_{\theta}(\hat{\mathbf{x}}_i), \mathbf{c}_j)}
\end{equation}
The unlabeled sample $\hat{\mathbf{x}}_{i}$ will be then integrated into the class-specific cluster with the highest probability. We can further adjust the class centroid $\mathbf{c}_k$ considering the unlabeled samples: 
\begin{equation}
    \mathbf{c}_k = \frac{N_k}{N_k + \hat{N}_k} \sum_{i=1}^{N_k} W_{k,i} \cdot \mathbf{h}_i^{k} + \frac{\hat{N}_k}{N_k + \hat{N}_k} \sum_{i=1}^{\hat{N}_k} \hat{p}_{k,i} \cdot \hat{\mathbf{h}}_i^{k}
\end{equation}
where $\hat{\mathbf{h}}_{i}^{k} = f_{\theta}(\hat{\mathbf{x}}_{i}^{k})$, $\hat{N}_k$ is the number of samples of class $k$ in $\mathcal{D}_{u}$ with the propagated label.

The class centroids are initialized and adjusted by both labeled and unlabeled samples, from which we formalize the \textit{regularization loss} derived from the labeled samples: 
\begin{equation}
    L_{Reg}(\theta) = -\sum\nolimits_{k} log W_{\theta}(y=k | \mathbf{x})
\end{equation}

As both the reconstruction and regularization losses are normalized, we define the global optimization objective as:
\begin{equation}
    \min\nolimits_{\theta} (L_R + \lambda L_{Reg})
\end{equation}
where $\lambda \geq 0$ is a hyperparameter that balances the two losses. Importantly, $L_{Reg}$ is included such that the embedding process not only serves to reduce the dimensions -- it is actively conditioned to facilitate the encoder in learning class-separable embeddings. In practice, SMATE is not sensitive to $\lambda$; then for all the experiments, we set $\lambda$ to 1.

\section{Experiments}
\subsection{Experimental setup}
\vspace{-0.5em}
The model\footnote{The source code: https://github.com/JingweiZuo/SMATE} was trained using the Adam optimizer on a single Tesla V100 GPU of 32 Go memory with CUDA 10.2.
We train an SVM classifier with radial basis function kernel on the learned embedding space, which is evaluated on the newly released UEA archive \cite{Hoang2018The2018}\footnote{The datasets can be found in www.timeseriesclassification.com}
for supervised analysis, where the datasets \{\textit{ArticularyWordR., Epilepsy, Heartbeat, SelfRegulationSCP1}\} from four different domains are adopted for semi-supervised study. 
\vspace{-0.5em}
\subsection{Classification Performance Evaluation}
We use the accuracy as the default metric \cite{Bagnall2020TheAdvances} for the supervised tasks. We also report the number of Win/Ties and the average rank \cite{ZhangTapNet:Network} of different methods. 
\subsubsection{Comparison Methods}
\begin{itemize}
    \item Distance-based Nearest Neighbor (1NN) on non-normalized (\textit{non-norm}) or normalized (\textit{norm}) MTS \cite{Shokoohi-Yekta2017GeneralizingAccess}: $\textbf{1NN-ED}$; $\textbf{1NN-DTW}_I$; $\textbf{1NN-DTW}_D$; $\textbf{1NN-DTW}_A$. 
    
    \item  Bag-of-patterns classifier. $\textbf{WEASEL+MUSE}$\cite{Schafer2017MultivariateWEASEL+MUSE}.
    
    \item Deep Learning-based classifier. $\textbf{USRL}$ \cite{Franceschi2019UnsupervisedSeries}; $\textbf{TapNet}$ \cite{ZhangTapNet:Network}; $\textbf{MLSTM-FCN}$ \cite{KarimMultivariateClassification}; $\textbf{CA-SFCN}$ \cite{Hao2020ASeries}; $\mathbf{SMATE_{NR}}$: SMATE without supervised Regularization, instead, a \textit{Softmax} layer is applied on the embedding.
\end{itemize}

\subsubsection{Results Analysis}
\begin{table*}[htbp]
\centering
\vspace{-0.5em}
\caption{\label{PerformanceComparison}Performance Comparison for MTS classification over UEA MTS archive}
\scalebox{0.56}{
\begin{tabular}{|m{3cm}<{\centering} |m{1.2cm}<{\centering} m{1.2cm}<{\centering} m{1.2cm}<{\centering} m{1.2cm}<{\centering} m{1.2cm}<{\centering} m{1.2cm}<{\centering} m{1.5cm}<{\centering} m{1.2cm}<{\centering} m{1.2cm}<{\centering} m{1.2cm}<{\centering} m{1.5cm}<{\centering} m{1.5cm}<{\centering} m{1.6cm}<{\centering} m{1.5cm}<{\centering}|}

\hline

Dataset               & SMATE          & SMATE$_{\text{NR}}$ & USRL                    & TapNet         & MLSTM -FCN     & CA-SFCN        & WEASEL +MUSE   & 1NN-ED & 1NN-DTW$_{I}$  & 1NN-DTW$_{D}$  & 1NN-ED (\textit{norm}) & 1NN-DTW$_{I}$ (\textit{norm}) & 1NN-DTW$_{D}$ (\textit{norm}) & 1NN-DTW$_{A}$ (\textit{norm})  \\ 
\hline
\hline
ArticularyWordR.      & \textbf{0.993} & 0.987               & 0.987                   & 0.987          & 0.973          & 0.97           & 0.99           & 0.97   & 0.98           & 0.987          & 0.97                   & 0.98                          & 0.987                         & 0.987                          \\
AtrialFibrillation    & 0.133          & 0.133               & 0.133                   & \textbf{0.333} & 0.267          & \textbf{0.333} & \textbf{0.333} & 0.267  & 0.267          & 0.2            & 0.267                  & 0.267                         & 0.22                          & 0.267                          \\
BasicMotions          & \textbf{1}     & \textbf{1}          & \textbf{1}              & \textbf{1}     & 0.95           & \textbf{1}     & \textbf{1}     & 0.675  & \textbf{1}     & 0.975          & 0.676                  & \textbf{1}                    & 0.975                         & \textbf{1}                     \\
CharacterTrajectories & 0.984          & \textbf{0.997}      & 0.994                   & \textbf{0.997} & 0.985          & 0.988          & 0.99           & 0.964  & 0.969          & 0.99           & 0.964                  & 0.969                         & 0.989                         & 0.989                          \\
Cricket               & 0.986          & 0.968               & 0.986                   & 0.958          & 0.917          & 0.972          & \textbf{1}     & 0.944  & 0.986          & \textbf{1}     & 0.944                  & 0.986                         & \textbf{1}                    & \textbf{1}                     \\
DuckDuckGeese         & N/A            & N/A                 & \textbf{\textbf{0.675}} & 0.575          & \textbf{0.675} & N/A            & 0.575          & 0.275  & 0.55           & 0.6            & 0.275                  & 0.55                          & 0.6                           & 0.567                          \\
EigenWorms            & N/A            & N/A                 & 0.878                   & 0.489          & 0.504          & N/A            & \textbf{0.89}  & 0.55   & 0.603          & 0.618          & 0.549                  & N/A                           & 0.619                         & N/A                            \\
Epilepsy              & 0.964          & 0.946               & 0.957                   & 0.971          & 0.761          & 0.986          & \textbf{1}     & 0.667  & 0.978          & 0.964          & 0.666                  & 0.978                         & 0.964                         & 0.979                          \\
ERing                 & \textbf{0.981} & 0.904               & 0.88                    & 0.904          & 0.941          & 0.856          & 0.964          & 0.93   & 0.93           & 0.93           & 0.93                   & 0.93                          & 0.93                          & 0.93                           \\
EthanolConcentration  & 0.399          & 0.373               & 0.236                   & 0.323          & 0.373          & 0.323          & \textbf{0.43}  & 0.293  & 0.304          & 0.323          & 0.293                  & N/A                           & 0.323                         & 0.316                          \\
FaceDetection         & \textbf{0.647} & 0.556               & 0.528                   & 0.556          & 0.545          & N/A            & 0.545          & 0.519  & 0.513          & 0.529          & 0.519                  & 0.5                           & 0.529                         & 0.529                          \\
FingerMovements       & \textbf{0.62}  & 0.55                & 0.54                    & 0.53           & 0.58           & 0.59           & 0.49           & 0.55   & 0.52           & 0.53           & 0.55                   & 0.52                          & 0.53                          & 0.509                          \\
HandMovementD.        & \textbf{0.554} & 0.365               & 0.27                    & 0.378          & 0.365          & 0.324          & 0.365          & 0.279  & 0.306          & 0.231          & 0.278                  & 0.306                         & 0.231                         & 0.224                          \\
Handwriting           & 0.421          & 0.335               & 0.533                   & 0.357          & 0.286          & 0.322          & 0.605          & 0.371  & 0.509          & \textbf{0.607} & 0.2                    & 0.316                         & 0.286                         & 0.601                          \\
Heartbeat             & 0.741          & 0.615               & 0.737                   & 0.751          & 0.663          & \textbf{0.756} & 0.727          & 0.62   & 0.659          & 0.717          & 0.619                  & 0.658                         & 0.717                         & 0.571                          \\
InsectWingbeat        & N/A            & N/A                 & 0.16                    & \textbf{0.208} & 0.167          & N/A            & N/A            & 0.128  & N/A            & 0.115          & 0.128                  & N/A                           & N/A                           & N/A                            \\
JapaneseVowels        & 0.965          & 0.924               & \textbf{0.989}          & 0.965          & 0.976          & 0.973          & 0.973          & 0.924  & 0.959          & 0.949          & 0.924                  & 0.959                         & 0.949                         & 0.959                          \\
Libras                & 0.849          & 0.834               & 0.867                   & 0.85           & 0.856          & 0.89           & 0.878          & 0.833  & \textbf{0.894} & 0.872          & 0.833                  & \textbf{0.894}                & 0.87                          & 0.879                          \\
LSST                  & 0.582          & 0.568               & 0.558                   & 0.568          & 0.373          & \textbf{0.674} & 0.59           & 0.456  & 0.575          & 0.551          & 0.456                  & 0.575                         & 0.551                         & 0.551                          \\
MotorImagery          & \textbf{0.59}  & \textbf{0.59}       & 0.54                    & \textbf{0.59}  & 0.51           & N/A            & 0.51           & 0.39   & N/A            & 0.5            & 0.51                   & N/A                           & 0.5                           & 0.5                            \\
N/ATOPS               & 0.922          & 0.87                & 0.944                   & 0.939          & 0.889          & \textbf{0.956} & 0.87           & 0.86   & 0.85           & 0.883          & 0.85                   & 0.85                          & 0.883                         & 0.883                          \\
PEMS-SF               & \textbf{0.803} & 0.744               & 0.688                   & 0.751          & 0.699          & N/A            & N/A            & 0.705  & 0.734          & 0.711          & 0.705                  & 0.734                         & 0.711                         & 0.73                           \\
PenDigits             & 0.98           & 0.98                & \textbf{0.983}          & 0.98           & 0.978          & 0.975          & 0.948          & 0.973  & 0.939          & 0.977          & 0.973                  & 0.939                         & 0.977                         & 0.977                          \\
Phoneme               & 0.177          & 0.19                & \textbf{0.246}          & 0.175          & 0.11           & 0.19           & 0.19           & 0.104  & 0.151          & 0.151          & 0.104                  & 0.151                         & 0.151                         & 0.151                          \\
RacketSports          & 0.849          & 0.816               & 0.862                   & 0.868          & 0.803          & 0.875          & \textbf{0.934} & 0.868  & 0.842          & 0.803          & 0.868                  & 0.842                         & 0.803                         & 0.858                          \\
SelfRegulationSCP1    & \textbf{0.887} & 0.874               & 0.771                   & 0.739          & 0.874          & 0.734          & 0.71           & 0.771  & 0.765          & 0.775          & 0.771                  & 0.765                         & 0.775                         & 0.786                          \\
SelfRegulationSCP2    & \textbf{0.567} & 0.539               & \textbf{0.556}          & 0.55           & 0.472          & N/A            & 0.46           & 0.483  & 0.533          & 0.539          & 0.483                  & 0.533                         & 0.539                         & 0.539                          \\
SpokenArabicDigits    & 0.979          & 0.967               & 0.956                   & 0.983          & \textbf{0.99}  & 0.982          & 0.982          & 0.967  & 0.96           & 0.963          & 0.967                  & 0.959                         & 0.963                         & 0.963                          \\
StandWalkJump         & \textbf{0.533} & 0.4                 & 0.4                     & 0.4            & 0.067          & 0.2            & 0.333          & 0.2    & 0.333          & 0.2            & 0.2                    & 0.333                         & 0.2                           & 0.333                          \\
UWaveGestureLibrary   & 0.897          & 0.869               & 0.884                   & 0.894          & 0.891          & 0.8            & \textbf{0.916} & 0.881  & 0.868          & 0.903          & 0.81                   & 0.868                         & 0.903                         & 0.9                            \\ 
\hline
\hline
Avg. Rank             & 3.85           & 6.19                & 5.9                     & 4.73           & 7.33           & 5.45           & 4.66           & 9.3    & 7.43           & 6.37           & 9.37                   & 7.88                          & 6.83                          & 6.21                           \\
Wins (Ties)~          & \textbf{11}    & \textbf{3}          & 6                       & 5              & 2              & 5              & 8              & 0      & 2              & 2              & 0                      & 2                             & 1                             & 2                              \\

\hline
\end{tabular}}
\vspace{-2.5em}
\end{table*}
Table \ref{PerformanceComparison} shows the results comparison with the baselines. ``N/A'' indicates the model is not applicable due to memory overflow. Overall, SMATE defends its reliability with 11 Wins/Ties and the highest average rank of 3.85 among all the baselines. The current state-of-the-art deep learning methods (TapNet, CA-SFCN) and the powerful data mining method (WEASEL+MUSE) have close ranks (4.73/5.45/4.66). Besides, USRL and SMATE$_\text{{NR}}$ perform much worse than SMATE with the same SVM classifier, confirming the reliability of our supervised regularization on the embedding space. Moreover, SMATE achieves the best performance among the baselines on all the datasets of EEG/MEG applications \cite{Hoang2018The2018} (\textit{FaceDetection}, \textit{FingerMovements}, \textit{HandMovementDirection}, \textit{MotorImagery}, \textit{SelfRegulationSCP1}, \textit{SelfRegulationSCP2}), where the signals (i.e., variables) generally have strong and dynamic dependencies with each other. The spatial dynamic interactions could be essential characteristics that SMATE has successfully captured.

\vspace{-0.5em}
\subsection{Semi-supervised Classification Performance}
For semi-supervised tasks, we evaluate the classifier's accuracy at different supervision levels.
For comparison, we applied one classic model \textbf{1NN-DTW-D} \cite{Chen2013DTW-D:Example} and three recently proposed semi-supervised deep learning models: \textbf{USRL} \cite{Franceschi2019UnsupervisedSeries}, \textbf{Semi-TapNet} \cite{ZhangTapNet:Network} and \textbf{MTL} \cite{JawedSelf-SupervisedClassification}. Since \textbf{1NN-DTW-D} and \textbf{MTL} are initially designed for UTS, we adapt them by:
\begin{itemize}
    \item Adopting \textit{$DTW_{D}$} \cite{Shokoohi-Yekta2017GeneralizingAccess} as distance in \textbf{1NN-DTW-D}.

    \item Updating the MTS network optimization metrics in \textbf{MTL}.
\end{itemize}

\begin{figure*}[htbp]
\centering 
\subfloat[ArticularyWordR. (Motion)]{
    \includegraphics[width=0.21\linewidth, height=2.2cm]{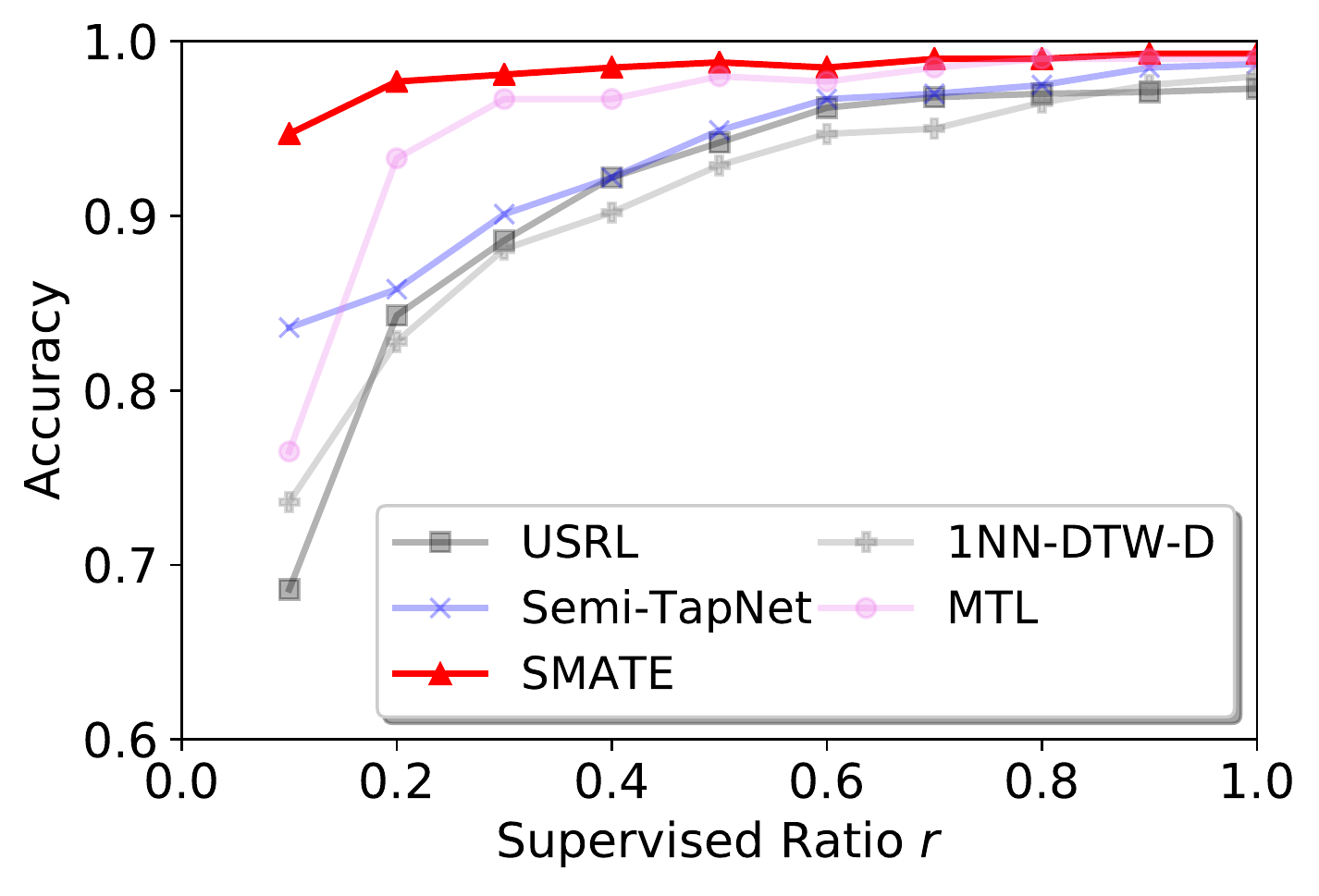}
    \label{sslfig:subfig1}
}
\centering
\subfloat[Epilepsy (Human Activity)]{
    \includegraphics[width=0.21\linewidth, height=2.2cm]{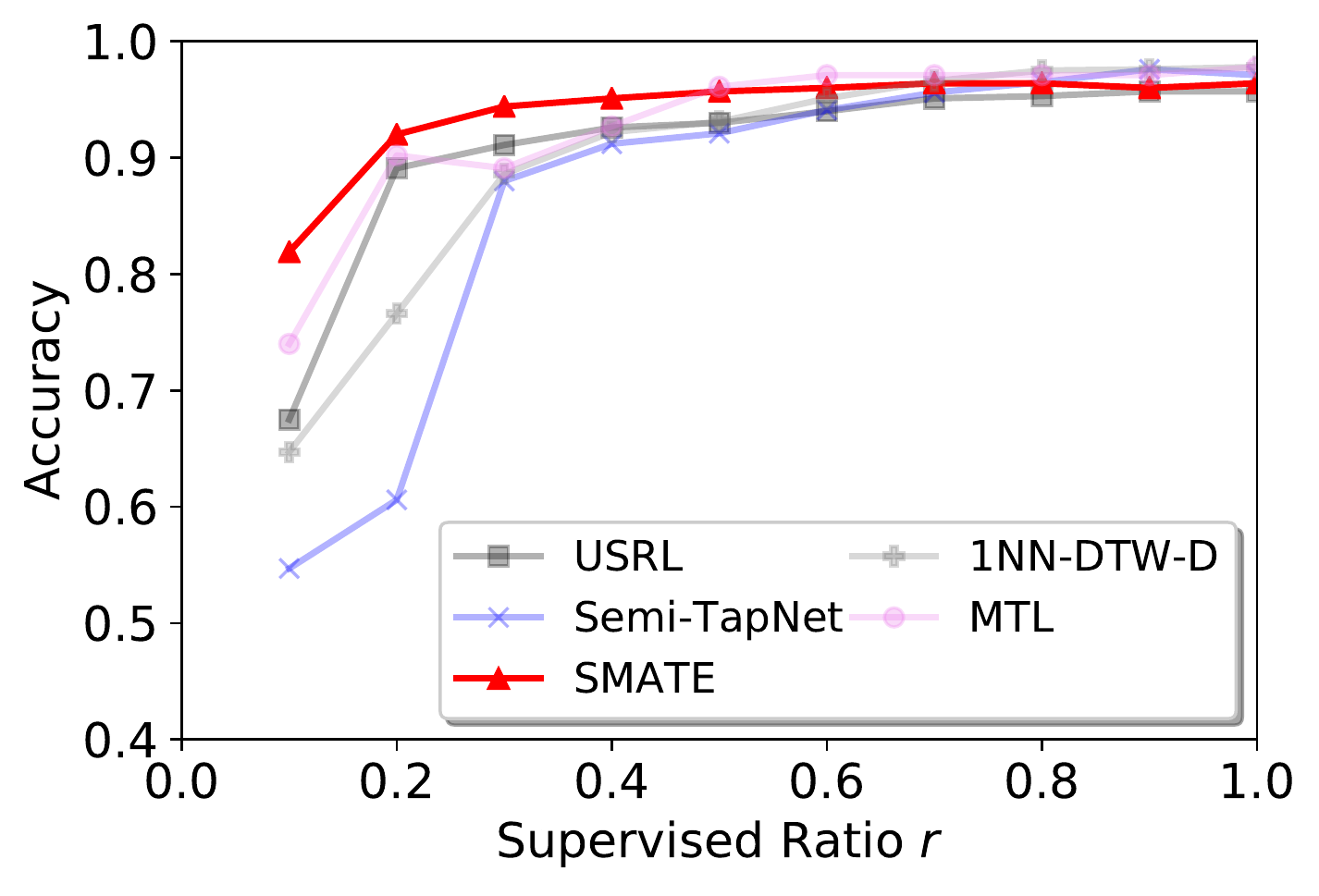}
    \label{sslfig:subfig2}
}
\centering 
\subfloat[Heartbeat (Audio Spectra)]{
    \includegraphics[width=0.21\linewidth, height=2.2cm]{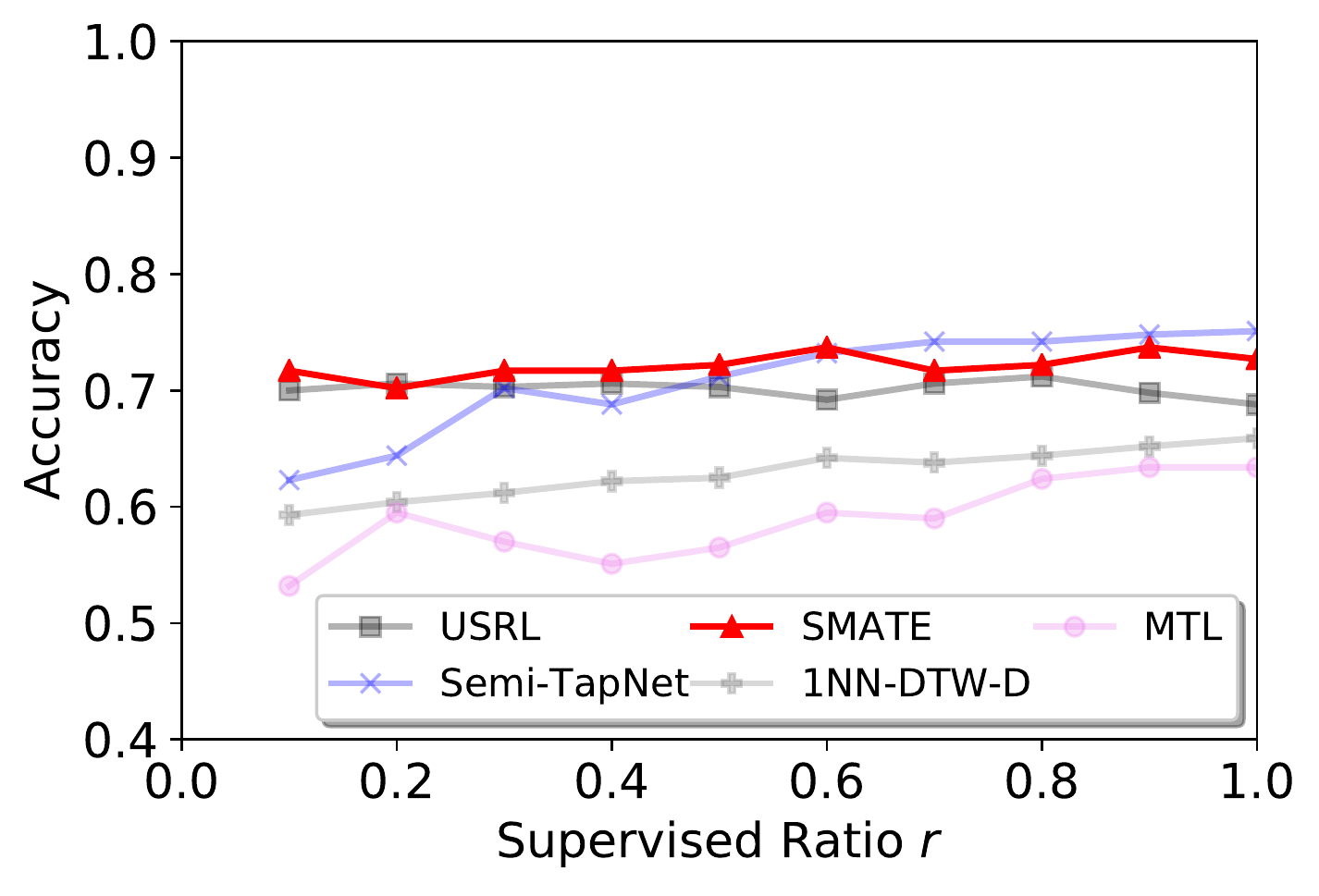}
    \label{sslfig:subfig3}
}
\centering 
\subfloat[SelfRegulationSCP1 (EEG/MEG)]{
    \includegraphics[width=0.22\linewidth, height=2.2cm]{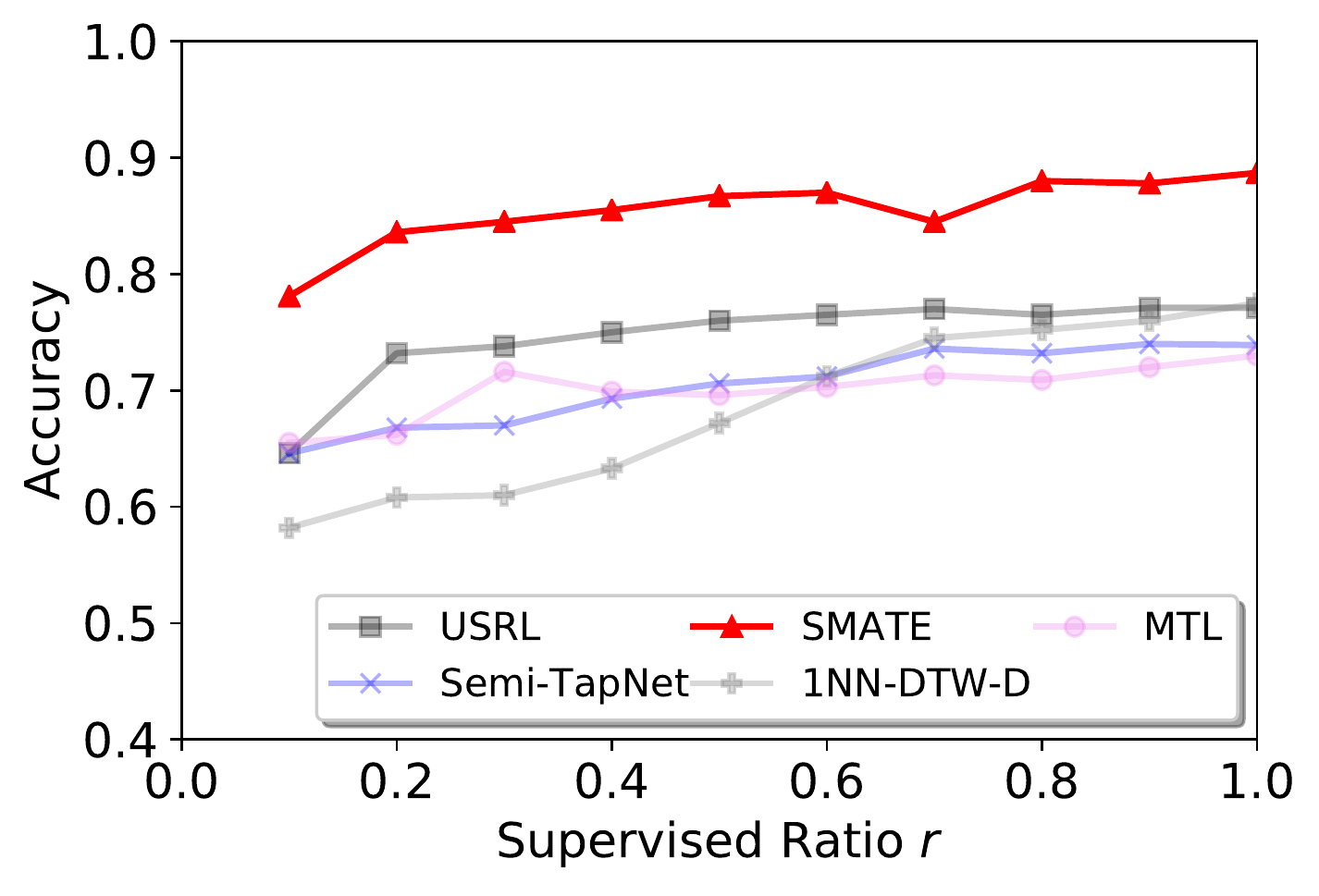}
    \label{sslfig:subfig4}
}

\hfill
\vspace{-0.5em}
\caption{Semi-supervised performance comparison on datasets from different domains}
\label{PerformanceComparisonSSL}
\vspace{-1em}
\end{figure*}

The results in Figure \ref{PerformanceComparisonSSL} shows the superiority of SMATE over the baselines. In a Motion Recognition task (\textit{ArticularyWorkR.}), from 10\% labeled training set to fully labeled one, the accuracy of SMATE varies only by 0.046, compared to INN-DTW-D (0.264), USRL (0.286), Semi-TapNet (0.151) and MTL(0.225), showing that SMATE is capable of learning a class-separable representation under weak supervision. This conclusion is also demonstrated in EEG/MEG applications (\textit{SelfRegulationSCP1}), with 10\% labeled samples, SMATE obtained a higher accuracy (0.781) than fully supervised 1NN-DTW-D (0.775), USRL (0.771), Semi-TapNet (0.739) and MTL (0.730). In an Audio Spectra task (\textit{Heartbeat}), though the fully supervised accuracy of SMATE (0.741) is not as good as Semi-TapNet (0.751), the weakly supervised SMATE with 10\% labeled samples performs the best among all the models, indicating the reliability of the semi-supervised model. 

\subsection{Visualization \& Interpretation of the Representation Space}
Apart from the thorough exploration of the weakly labeled samples, the representation space learned via SMATE shows good interpretability compared to the traditional Deep Learning models \cite{ZhengLNCSNetworks,IJCAI1510710,Hao2020ASeries,KarimMultivariateClassification}. 

\begin{figure*}[htbp]
\vspace{-1em}
\centering
\subfloat[\centering Without any regularization]{%
    \includegraphics[width=0.22\linewidth, height=2.6cm]{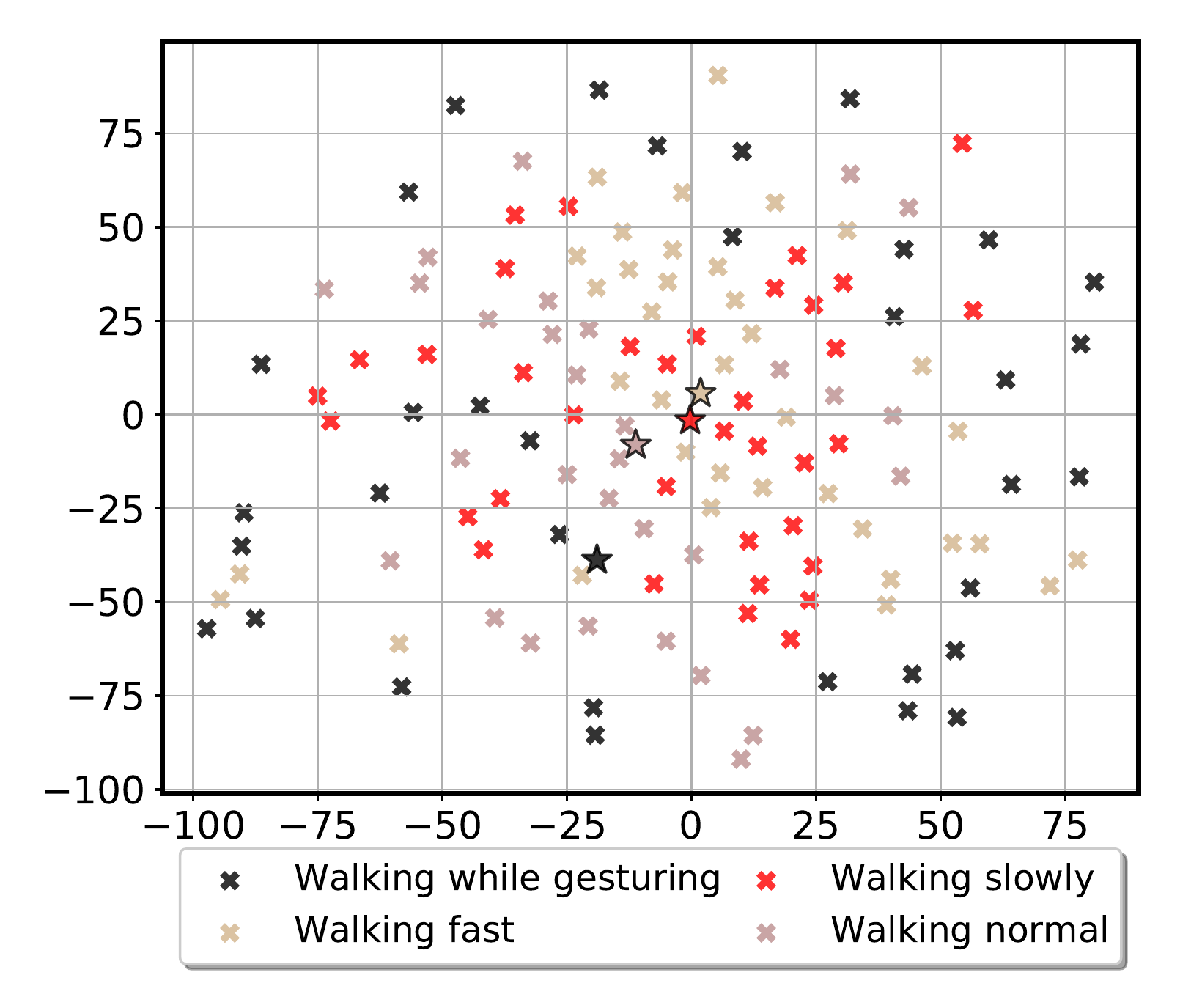}
    \label{fig:subfig1}
}
\subfloat[\centering Regularization step 1]{%
    \includegraphics[width=0.22\linewidth, height=2.6cm]{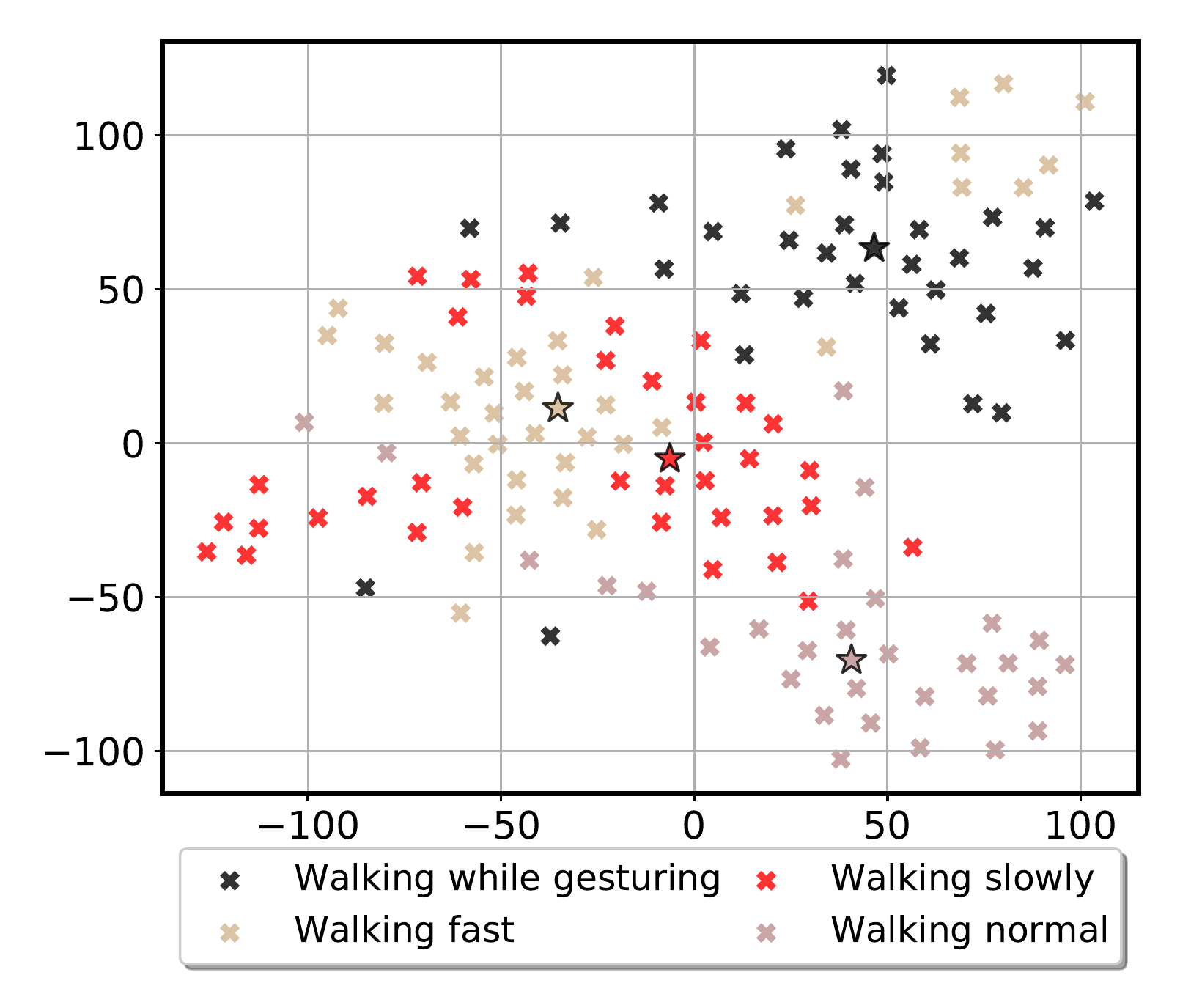}
    \label{fig:subfig2}
}
\subfloat[\centering Regularization step 2]{%
    \includegraphics[width=0.22\linewidth, height=2.6cm]{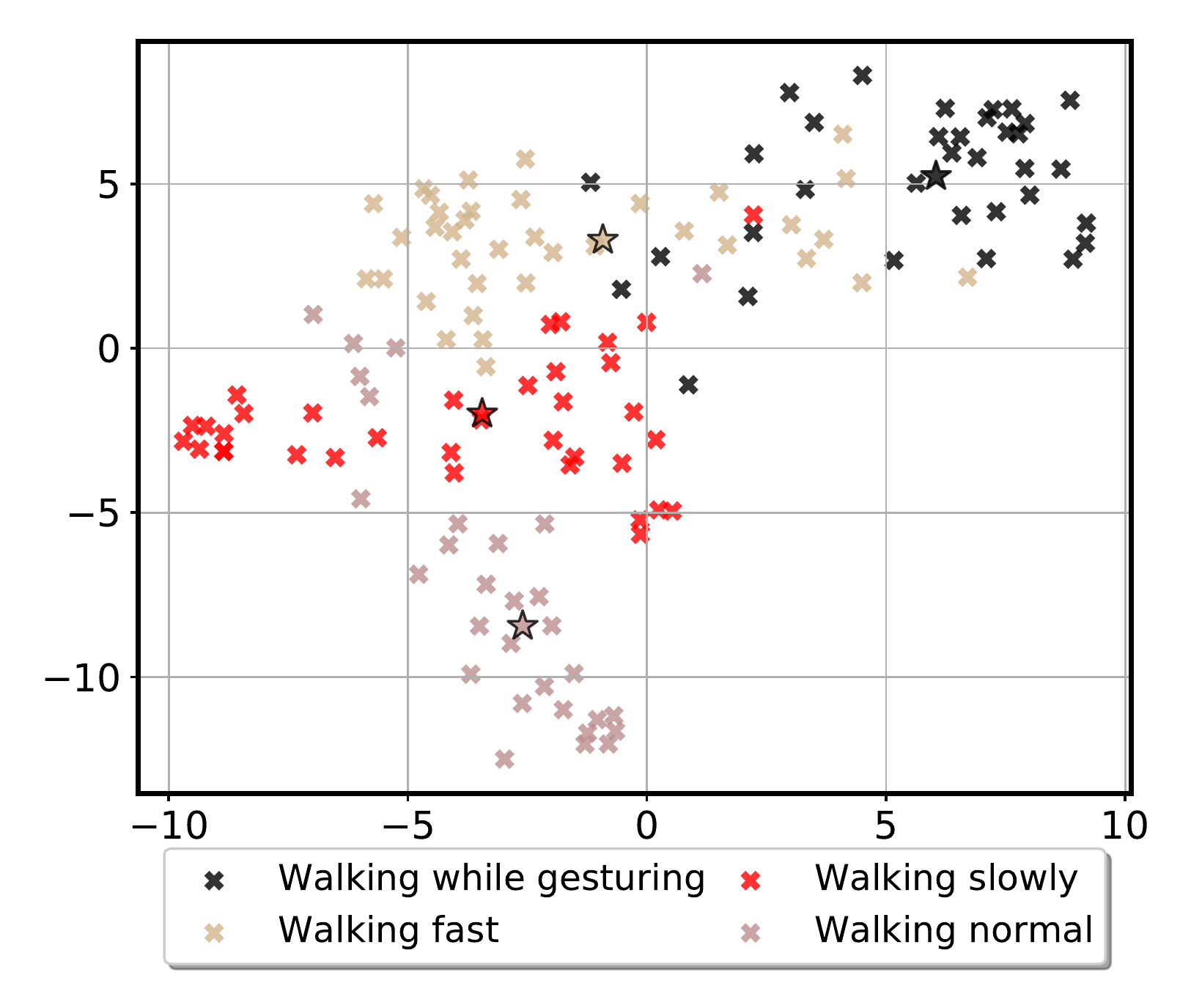}
    \label{fig:subfig3}
}
\subfloat[\centering Regularization step 3]{%
    \includegraphics[width=0.22\linewidth, height=2.6cm]{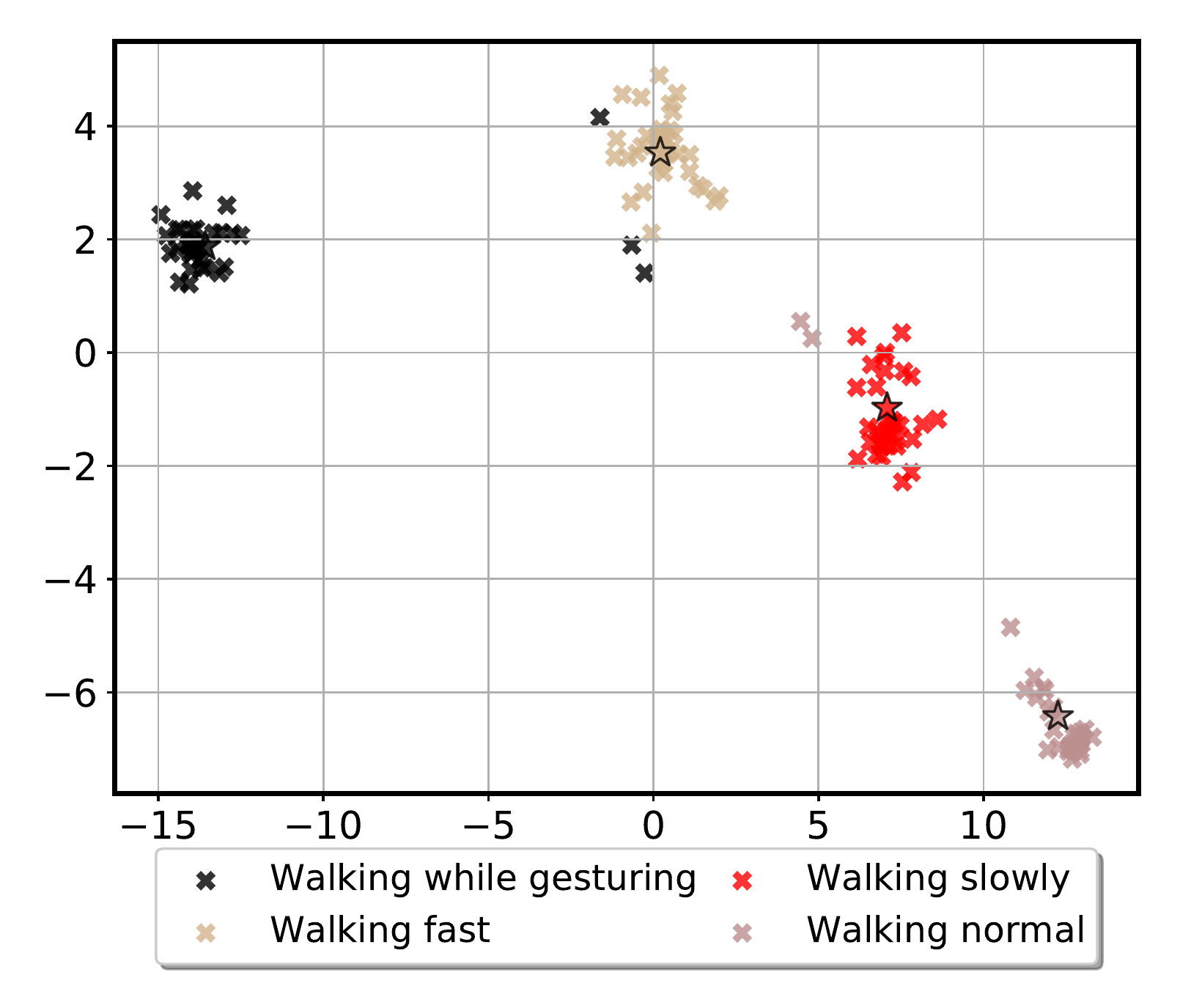}
    \label{fig:subfig4}
}
\vspace{-0.5em}
\caption{The t-SNE visualization of the representation space for the \textit{Epilepsy} dataset. We set the supervised ratio to 0.1. The colors of the embeddings represent their inherent labels, which are not fully adopted for training. The class centroids are marked by $\star$.}
\label{t-NSE_visualization}
\vspace{-1em}
\end{figure*}

We show in Figure \ref{t-NSE_visualization} that t-SNE visualization of the representation space for the \textit{Epilepsy} dataset, which contains four human activities: \textit{Walking while gesturing}, \textit{Walking slowly}, \textit{Walking fast}, \textit{Walking normally}. The 137 samples in the training set are projected to the representation space with 10\% labels adopted for training. We show respectively the space visualization at each regularization step. The results suggest that the representation space is interpretable for not only the effect of the weak supervision at each step but also the classification results. For instance, in Figure \ref{fig:subfig4}, three samples of \textit{Walking while gesturing} stay close to the class centroid of \textit{Walking fast}, which may lead to the misclassification of certain samples in the two classes. To improve the classifier, more labels of the two classes in the training set can be added.

\subsection{Performance of Spatial Modeling Block (SMB)}
To validate the Spatial Modeling Block (SMB), firstly, we compare the classification accuracy of SMATE with or without integrating SMB on the 27 datasets that SMATE has successfully executed. Then we rebuild SMATE by replacing SMB with the following modules in the state-of-the-art work which learn the variable relationships of MTS: \textbf{Random Dimension Permutation (RDP)} in TapNet \cite{ZhangTapNet:Network} and \textbf{Squeez-and-Excitation (SE)} in MLSTM-FCN \cite{KarimMultivariateClassification}. Briefly, SMATE-SMB achieves [17 Wins$\mid$8 Ties$\mid$2 Losses] to SMATE-NonSMB, indicating that SMB contributes to a better MTS representation. Besides, SMB performs better than other modules: [14 Wins$\mid$8 Ties$\mid$5 Losses] to SE, [12 Wins$\mid$9 Ties$\mid$6 Losses] to RDP. RDP performs relatively better than SE, as a set of grouped variables produced by RDP provides various MTS views, allowing exploring the interactions between the subsets of all variables more thoroughly. However, extra parameters for variable groups are introduced. SE is a parameter-free module but considers each variable has a unique and stable state when interacting with others, which ignores the dynamic features in time series. SMB answers both the questions of the parameter-free settings and the dynamic interactions. The results show that capturing the spatial dynamic interactions at the sub-sequence level performs better than modeling the variable interactions at the sequence level \cite{ZhangTapNet:Network,KarimMultivariateClassification}.
\vspace{-0.5em}
\section{Conclusion}
\vspace{-0.5em}
In this paper, we proposed SMATE, to learn the Spatial-temporal representation on weakly-labeled multivariate time series. The weak supervision on the embedding space allows building a reliable classifier, which is extremely valuable in real-life scenarios with label shortage issues. 
The results show that the evolving variable interactions (i.e., \textit{spatial dynamics}) play an essential role in modeling multivariate time series. 
Moreover, SMATE allows for visual interpretability in both the learned representation and the semi-supervised representation learning process.
Our future work will be oriented towards extending SMATE to support multivariate time series with missing values and unequal length in more realistic scenarios.
\vspace{-1.5em}
\section*{Acknowledgements}
\vspace{-0.5em}
This research was supported by DATAIA convergence institute as part of the \textit{Programme d’Investissement d’Avenir} (ANR-17-CONV-0003) operated by DAVID Lab, UVSQ, Université Paris-Saclay. The authors would like to thank Anthony Bagnall and his team for providing the community with valuable datasets and source codes in the UEA \& UCR Time Series Classification Repository. The authors would also like to thank Eamonn Keogh for his careful review and remarks on the preliminary version of this paper, as well as Nicoleta Preda and Zaineb Chelly for their suggestions.

\vspace{-1em}
\bibliographystyle{IEEEtran}
\bibliography{references.bib,refComplement.bib}
\end{document}